\documentclass[journal]{IEEEtran}
\usepackage{cite}
\usepackage{amsmath,amssymb,amsfonts}
\usepackage{algorithmic}
\usepackage{graphicx}
\usepackage{textcomp}
\usepackage[caption=false]{subfig}
\usepackage{relsize}

\def\BibTeX{{\rm B\kern-.05em{\sc i\kern-.025em b}\kern-.08em
    T\kern-.1667em\lower.7ex\hbox{E}\kern-.125emX}}

\providecommand{\norm}[1]{\lVert#1\rVert_2}

\newcommand*{\defeq}{\stackrel{\mathsmaller{\mathsf{def}}}{=}} 
    
\begin{document}

\title{Robust Re-identification of Manta Rays from Natural Markings by Learning Pose Invariant Embeddings}

\author{Olga~Moskvyak, Frederic~Maire, Asia~O.~Armstrong, Feras~Dayoub and~Mahsa~Baktashmotlagh
\thanks{O. Moskvyak, F. Maire and F. Dayoub are with the School of Electrical Engineering 
and Computer Science, Queensland University of Technology, Brisbane, QLD 4000, Australia.}
\thanks{A.~O.~Armstrong is with the School of Biomedical Science, The University of Queensland, St Lucia, QLD 4072, Australia.}%
\thanks{M.~Baktashmotlagh is with the School of Information Technology and Electrical Engineering, The University of Queensland, St Lucia, QLD 4072, Australia.}%
\thanks{Corresponding author O. Moskvyak: olga.moskvyak@hdr.qut.edu.au}}

\maketitle

\begin{abstract}
Visual identification of individual animals that bear unique natural body markings is an important task in wildlife conservation. The photo databases of animal markings grow larger and each new observation has to be matched against thousands of images. Existing photo-identification solutions have constraints on image quality and appearance of the pattern of interest in the image. These constraints limit the use of photos from citizen scientists. We present a novel system for visual re-identification based on unique natural markings that is robust to occlusions, viewpoint and illumination changes. We adapt methods developed for face re-identification and implement a deep convolutional neural network (CNN) to learn embeddings for images of natural markings.
The distance between the learned embedding points provides  a dissimilarity measure between the corresponding input images. 
The network is optimized using the triplet loss function and the online semi-hard triplet mining strategy. The proposed re-identification method is generic and not species specific. We evaluate the proposed system on image databases of manta ray belly patterns and humpback whale flukes.
To be of practical value and adopted by marine biologists, a re-identification system needs to have a top-10 accuracy of at least 95\%. 
The proposed system achieves this performance standard. 
\end{abstract}



\section{Introduction}
\label{sec:introduction}
\IEEEPARstart{R}{e-identification} of animal individuals by unique natural markings in photo databases is an effective and non-invasive mark-recapture tool for monitoring populations \cite{into-photo-id-rays-sharks}. 
Tracking population dynamics of animals such as manta rays is critical owing to their vulnerable conservation status, and economic importance in both ecotourism and fisheries \cite{review-manta-latest}.
These species cannot sustain heavy exploitation \cite{marine-croll}, and the trade of manta ray gill rakers is believed to be responsible for driving population declines upwards of 80\% in some locations \cite{marine-rohner}. Some species such as humpback whales are no longer threatened by commercial whaling. The conservation effort is now focused on the identification of individual humpback whales to better understand their use of breeding and feeding areas \cite{whales-conservation}.

\begin{figure}[!t]
    \centering
    \includegraphics[width=3.3in]{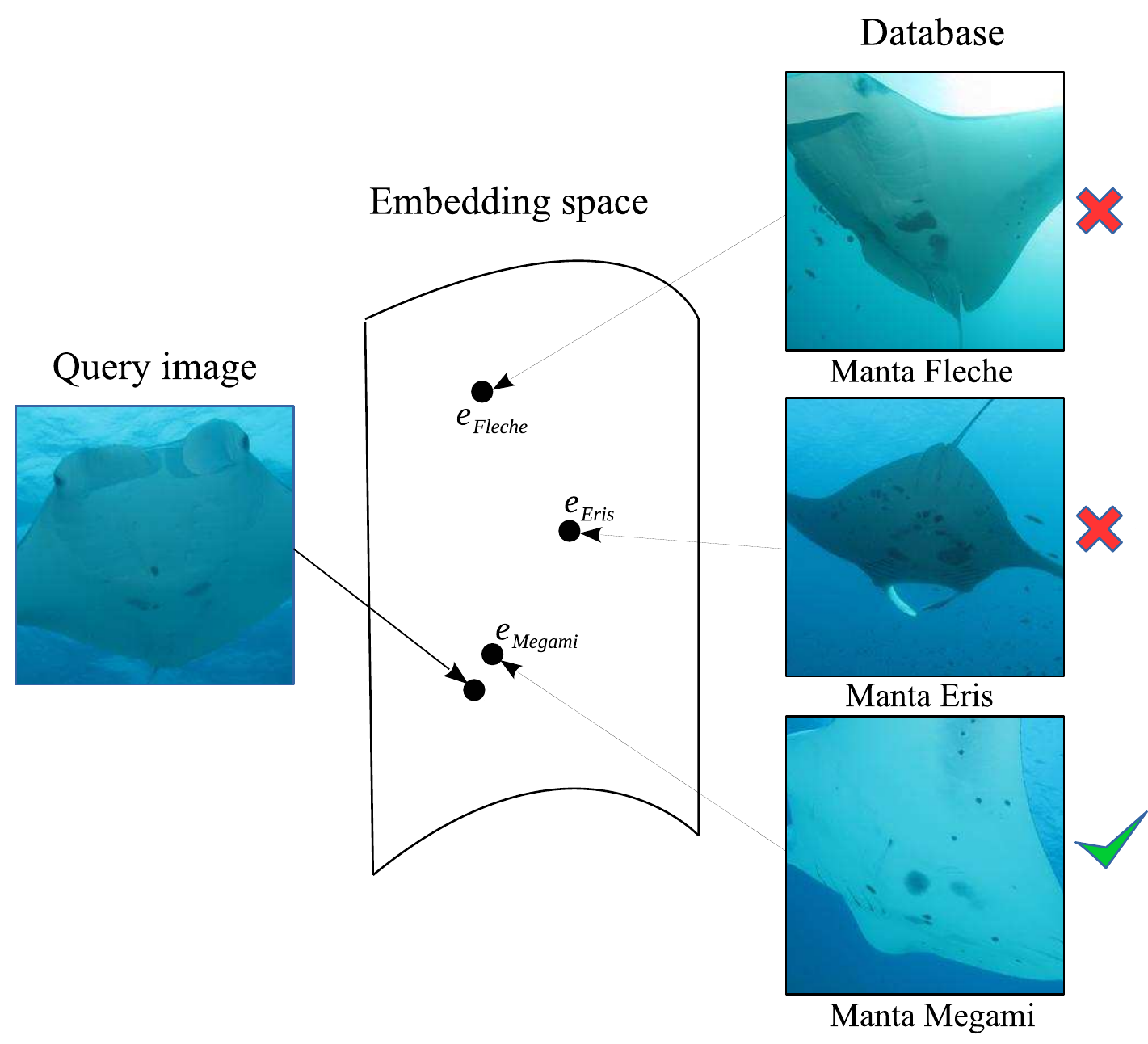}
    \caption{The proposed system learns embeddings for images from the database. The embeddings of the same individual are brought close together and embeddings of different manta rays are pushed further apart. A new query image is matched to the database by finding the closest points in the embedding space. The system learns embeddings that are invariant to viewing angle and illumination. Photo credit: Chris Garraway.}
    \label{fig:manta-intro}
\end{figure}

\begin{figure}[!t]
    \centering
    \includegraphics[width=3.3in]{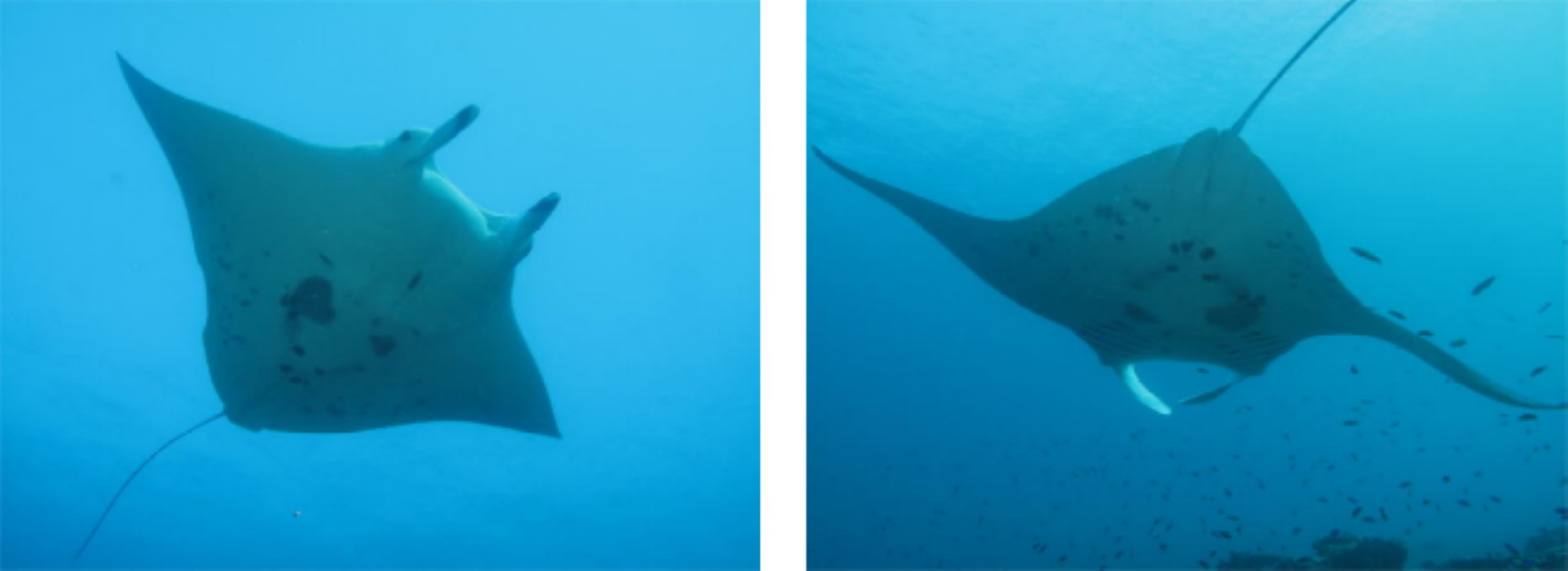}
    \caption{The camera angle can vary dramatically. Here the same manta ray, named Eris, was photographed from two different viewpoints. A homography transformation is required to align the belly patterns of the two images. Photo credit: Chris Garraway.}
    \label{fig:manta-example}
\end{figure}

Our research is focused on developing an automated system for visual re-identification of animals that bear unique natural 
markings. We demonstrate the suitability of the proposed system on photo databases of manta ray belly patterns and humpback whale flukes. Manta rays have a unique spot pattern on their ventral surface that allows individuals to be distinguished from one another. The spot pattern is conserved throughout the animals life, much like a human fingerprint. Examples of spot patterns are shown in Fig.~\ref{fig:manta-intro} and Fig.~\ref{fig:manta-example}.
Humpback whales have patterns of black and white pigmentation and scars on the underside of their tails that are unique to each whale. 

There are a number of factors that make animal re-identification 
based on natural markings challenging. 
Photo databases often rely on input from citizen scientists to fill in data gaps when researchers are not in the field. 
This means image quality cannot be guaranteed as camera parameters, and the angle of image capture vary. Other factors include poor visibility (especially for underwater images), illumination, and small objects occluding the pattern on the animal.


The current state-of-the-art manta ray recognition system Manta Matcher \cite{manta_matcher} requires the user to manually align and normalize the 2D orientation of the manta 
ray within the image, and select a rectangular region of interest containing the spot pattern. 
The Manta Matcher works best with photos taken perpendicular to the manta's ventral pattern with no reflective particles in the water and in good lighting conditions. 
In practice, these constraints limit the use of photos from citizen scientists and some marine biologists still do the identification manually using a handcrafted decision tree.
A common idea that has been applied to several species for recognizing individual animals is to search for an affine transformation matching patterns present in two distinct images  
(lizards \cite{review-aphis}, arthropods \cite{review-aphis-antropodes}, sharks \cite{review-sharks}, turtles \cite{ review-turtles}).
However, this approach requires annotating body landmarks on each individual image in the same order. 
This is not suitable for manta rays as we want to accept images of the animals in a wide range of poses  
with no requirement that all body landmarks are clearly visible. 

Convolutional neural networks (CNN) have been applied to the problem of animal identification as a classification problem \cite{review-pigs}, \cite{review-seals}, \cite{review-snolepard}, \cite{whale-kaggle}.
It means that the trained model is only able to identify the animals presented during training. 


It is highly desirable to have a system that is not only capable of recognizing animals whose images have been used to train the neural network, but also capable
of recognizing animals whose images have been added to the database well after the network has been trained without requiring the re-training of the network on these new instances.
This paper focuses on this more challenging and less studied problem for animal re-identification.





In this work we focus on eliminating some constraints 
of previous wildlife matching systems such as requirements for high image quality and a clear view of the animal markings in the image. We propose a solution inspired by advances in deep learning for face re-identification. Our approach uses a CNN to learn embeddings for images of animal markings in such a way that the distance between embeddings of the same individual is smaller than the distance between embeddings of this individual and other animals (see Fig.~\ref{fig:manta-intro}). 

The main contribution of this work is a novel visual wildlife re-identification system with the following properties:

\begin{enumerate}

    \item robustness to viewpoint changes, small occlusions and lighting conditions, and therefore ability to match images from citizen scientists;
    
    \item re-identification of individuals never seen  during training.
    

\end{enumerate}

The paper is organized as follows: in Section~\ref{sec:related-work} we discuss related work on re-identification. Our approach to learning embeddings is described in Section~\ref{sec:learning-embeddings}. The experimental setup and results are presented in Section~\ref{sec:experiments}.

\begin{figure*}[!t]
    \centering
    \includegraphics[width=0.95\textwidth]{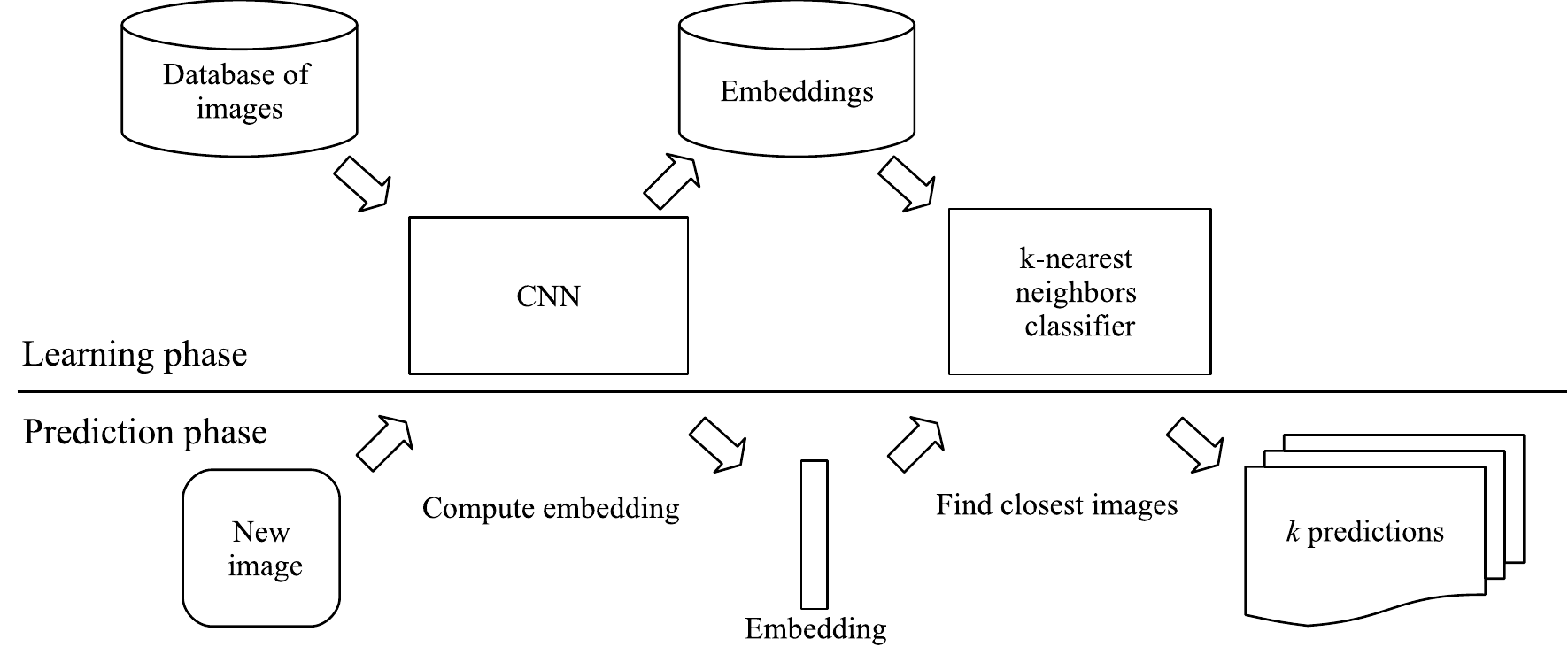}
    \caption{System architecture. All images from the current database are passed through the trained CNN model to compute embeddings and fit a nearest neighbors classifier in an embedding space. At the prediction step, $k$ predictions are obtained by computing an embedding using trained CNN and finding the closest points from the database using the nearest neighbors classifier.}
    \label{fig:train-predict}
\end{figure*}

\section{Related Work} \label{sec:related-work}


The techniques that have been proposed 
for photo-identification of animal natural markings vary in the core methods used, amount of user involvement and ability 
to be adapted to different species.
We review solutions used in practice for different cases.

Matching 
natural patterns has been approached by exhaustively generating two-dimensional affine transformations based on user provided key points and comparing each transformation of a candidate example with the examples stored in a repository \cite{review-aphis}, \cite{review-aphis-antropodes}, \cite{review-sharks}, \cite{review-turtles}. The algorithm was implemented in a solution called APHIS (Automated Photo-Identification Suite) and applied for re-identification of lizards \cite{review-aphis}, arthropods \cite{review-aphis-antropodes}, spotted raggedtooth sharks \cite{review-sharks} and turtle flippers \cite{review-turtles}. However, the method requires a user to select key points and identify the most distinctive spots for each image.

Some methods have been developed for specific species and, while performing well on these, are not easily transferable to other species. High-contrast colour patterns of humpback whale flukes \cite{review-whales} and dolphin dorsal fins \cite{review-dolphin} are matched by extracting hand-crafted features from corresponding segments obtained by overlaying a grid on a region of interest. This method is not robust to viewpoint changes.

Another approach identifies individual cetaceans from images showing the trailing edge of their fins by generating a representation of integral curvature of the nicks and notches along the trailing edge \cite{review-curvature}.

Current systems used in practice (Manta Matcher \cite{manta_matcher}, HotSpotter \cite{review-hotspotter}) are based on 
automated extraction and matching of keypoint features using the Scale-Invariant Feature Transform (SIFT) algorithm \cite{review-sift} with different modifications and enhancements to work on specific cases. While the algorithm works well on images that clearly show the pattern of interest, it is not robust to large changes in 
camera viewpoint, occlusions and variations in illumination.

The task of animal visual re-identification is related to the face recognition problem that has been extensively studied with deep learning in recent years \cite{parkhi2015deep, schroff2015facenet, sun2014deep}. The main idea is learning a function using a CNN that maps from a face image space to a space of a smaller dimension where the distance between the learned embedding vectors corresponds to a face similarity measure \cite{parkhi2015deep, chopra2005learning}. The network is trained on labelled image pairs or triplets to learn 
a face similarity measure under which the 
distance between the embeddings of faces from  the same person is reduced as much as possible and that of the distance between embeddings of faces of different people is increased. The problem is then reduced to the nearest neighbours search problem in Euclidean space, which can be solved by efficient approximate nearest neighbours search algorithms \cite{wang2014learning}.

The difference between face verification and animal re-identification is that a face image is typically normalized to an upright position whereas a pattern on an animal body is not necessarily in a canonical position 
and can appear at different angles.
See an example of the same manta ray 
viewed from different vantage points in Fig.~\ref{fig:manta-example}. A robust identification system should be invariant to the pose of the object of interest and viewing angle. In our previous work \cite{dicta-paper}, we investigated 
the difficulty of recognizing a set of artificially generated patterns subjected to various projective transformations 
to simulate the variations in appearance of natural markings from different vantage points. This previous  
study explored Siamese \cite{chopra2005learning} and Triplet \cite{wang2014learning} architectures with different loss functions for learning the homographic equivalence between patterns. It was concluded that these architectures with a relatively simple CNN in its core were suitable for pattern re-identification. The results were promising and we have now extended this approach to real images of animal markings in the wild.

\section{Learning embeddings} \label{sec:learning-embeddings}

Throughout the paper, we say that images from the same individual animal \textit{belong to the same class}. 
Images of different individual animals are  said to be from \textit{different classes}. The re-identification task can be formulated as a classification problem where the number of classes is in the order of thousands and not known in advance, and the number of examples for each class is small. 
The following section gives an overview of the architecture of our re-identification system.

\subsection{System architecture}

The system, illustrated in Fig.~\ref{fig:train-predict}, consists of a CNN that produces embeddings for images and a k-nearest neighbors classifier in the embedding space. 
During the learning phase, we train a CNN on a database of labelled images.
During the prediction phase, a new query image is fed to the network to produce an embedding.
The first $k$ animals in the embedding database that are closest to the embedding of the query image are returned.
Two outcomes are possible during the verification of the identity of the animals. 
Either the marine biologist confirmes that the query image corresponds to one of the $k$ returned animals or the query image is considered to be from a never seen before animal.  In the first case, the query image is added to the record of the recognized animal. In the second case, a new animal entry is created.
Over time, new images are added to the database but the CNN is not systematically retrained on the extended dataset.
The network is able to match against images that were in the database during training as well as against images added later.

\subsection{Model}


\begin{figure}[!t]
    \centering
    \includegraphics[width=3.3in]{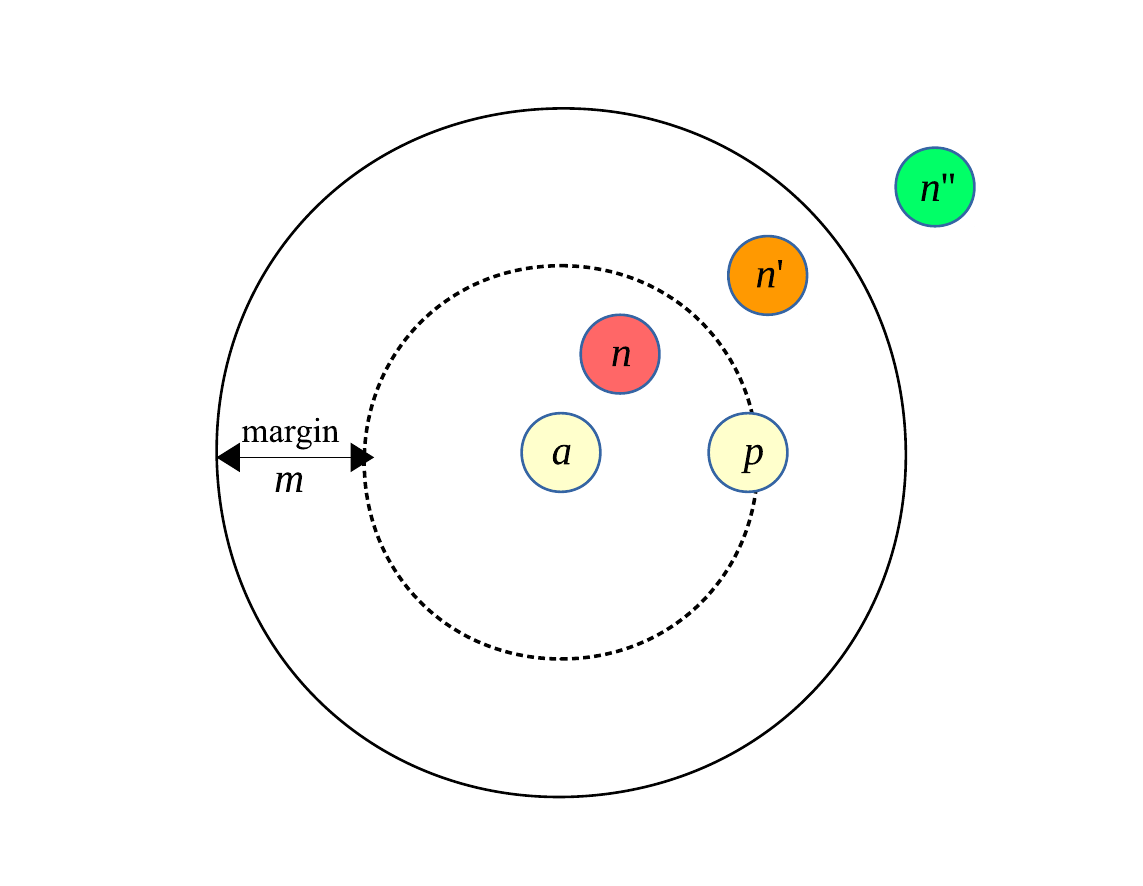}
    \caption{Examples of a hard triplet, a semi-hard triplet and an easy triplet of embeddings. Anchor embedding $a$, its closest positive point $p$ (same class as $a$) and the red negative $n$ (different class from $a$) form a hard triplet as the negative is closer to the anchor than the positive example. The triplet $(a, p, n^{\prime})$ is a semi-hard triplet as the orange negative $n^{\prime}$ lies within the margin from the positive. The triplet contributes a positive value to the loss function. Whereas the triplet $(a, p, n^{\prime\prime})$ with the green negative $n^{\prime\prime}$ is an easy triplet because it contributes zero to the loss function.}
    \label{fig:triplets-diagram}
\end{figure}

We have adapted a model proposed in FaceNet \cite{schroff2015facenet} that learns embeddings for faces by minimizing a triplet loss. Initially, it was claimed that representation learning with the triplet loss is inferior to a combination of classification  and verification losses \cite{triplet-classif1, triplet-classif2}. However, modifications of the triplet loss (angular loss \cite{angular-loss}, magnet loss \cite{magnet-loss}) and smart triplet mining strategies (semi-hard \cite{schroff2015facenet}, batch-hard \cite{DBLP:in-defence-triplet, 100k-iden}) has proved that a model can successfully learn an end-to-end mapping between images and an embedding space.

The model consists of convolutional layers to extract features from an input image, a global pooling layer over feature maps and a fully connected layer to produce an embedding vector. 
We compare different CNN architectures as a base network, see details in Section~\ref{sec:base-networks}. 
The convolutional layers output a 3D array (e.g. $ 8 \times 8 \times 512$) that is then passed to a global pooling layer.


A global pooling layer takes the average of each feature map along the spatial axes 
(e.g. a tensor $ 8 \times 8 \times 512$ is transformed into a tensor  $ 1 \times 512$). 
We follow FaceNet  \cite{schroff2015facenet} and favor a global average pooling layer instead of a fully connected layer after convolutional layers. The global pooling layer makes the output of the network invariant to the size of the input images. 
Moreover, as the layer has no parameters,  overfitting is avoided \cite{network-in-network}.
The layer sums out the spatial information so it is more robust to spatial transformations of its input. 
Pooled features maps are passed to a fully connected layer to produce an embedding vector.

\subsection{Loss function} 

Our model is optimized using the triplet loss function \cite{wang2014learning} which accepts triplets of images. Let us define a triplet $ (I^a, I^{+}, I^{-}) $ where an image $I^a$ (\textit{anchor}) and an image $I^{+}$ (\textit{positive}) are from the same class and an image $I^{-}$ (\textit{negative}) is from a different class. 
The function $ D $ between two input images $ I $ and  $ J $ is defined as the Euclidean distance between their embeddings $f(I)$ and $f(J)$. 
That is,  $ D(I, J) \defeq \norm{f(I) - f(J) }$

The triplet loss function $\mathcal{L}$ encourages the squared distances between positive pairs of embeddings 
to become smaller than the squared distances between negative pairs of embeddings by a given margin $ m $:

\begin{equation} \label{eq:triplet-loss}
    \mathcal{L} \defeq \sum_{i=1}^{N} \max (0, m + D(I^a_i, I_i^{+})^2 - D(I_i^{+}, I_i^{-})^2)
\end{equation}
where \( N \) is the number of training triplets.

We also did experiments with the Siamese network architecture \cite{chopra2005learning} and a contrastive loss function \cite{hadsell2006dimensionality} over randomly generated pairs, however, the results were not as good as the results obtained with the triplet loss function.

\begin{figure*}[!t]
    \centering
    \subfloat[Images of acceptable quality]{\includegraphics[width=0.45\textwidth]{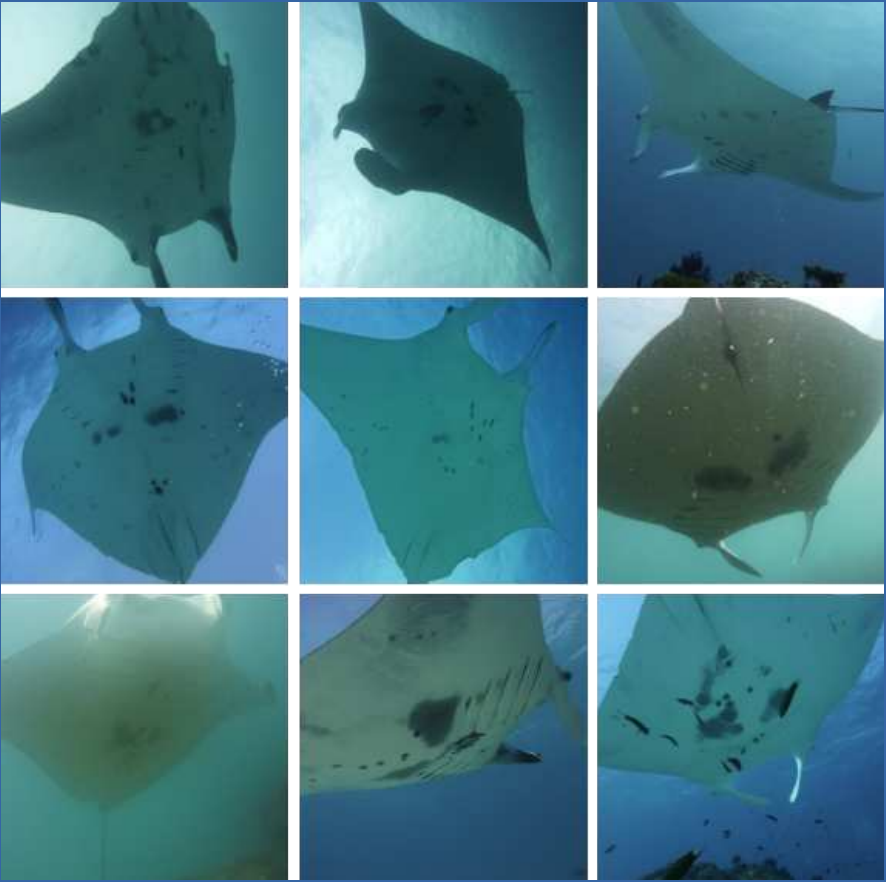}%
    \label{fig:manta-accepted}}
    \hfil
    \subfloat[Images excluded from the dataset]{\includegraphics[width=0.45\textwidth]{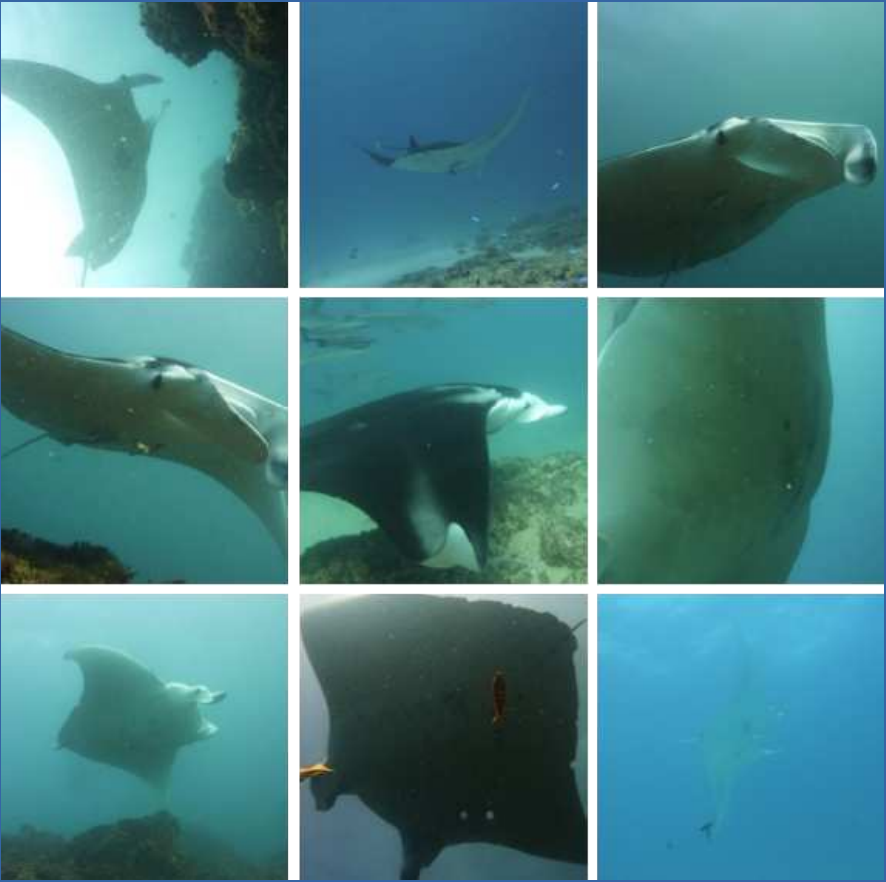}%
    \label{fig:manta-rejected}}
    \caption{Not all images of manta rays are acceptable for training and testing the system. 
    An image is accepted if the belly pattern is visible (even at oblique angles and in muddy water), see examples in (a). 
    We accept more challenging images than other methods \cite{manta_matcher}. Examples of excluded images in (b): back of the manta, side view, poor  underwater visibility. Photo credit: Lydie Couturier.}
    \label{fig:manta-dataset}
\end{figure*}

\subsection{Example mining} \label{sec:methodology-sample-mining}

The strategy for selecting triplets for learning embeddings plays an equal or more important role than the loss \cite{sampling-matters}. Generating random triplets for training with the triplet loss would result in many triplets that are already in a correct position and contribute zero loss to (\ref{eq:triplet-loss}). Several strategies have been proposed to optimize training with the triplet loss function. Batch-hard triplet mining \cite{DBLP:in-defence-triplet} selects the hardest positive (the furthest example from the same class) and the hardest negative (the closest example from a different class) within a batch for each anchor image. Another technique, distance-weighted sampling \cite{sampling-matters}, selects a negative example with a probability function of the distance to the negative example.

We follow the semi-hard triplet mining strategy proposed by \cite{schroff2015facenet} as we found experimentally that this approach works better than batch-hard strategy for our application domain. The triplet loss is calculated over triplets that contribute positive value to the loss function.
In other words, these negative examples lie within a margin from the positive examples (see Fig.~\ref{fig:triplets-diagram}). 
The selected triplets are not necessarily the hardest within a batch but they violate 
the constraint $ D(I_i^{a}, I_i^{+})^2 + m < D(I_i^{a}, I_i^{-})^2 $.

The triplet mining strategies listed above require computing embeddings in order to select triplets. 
This can be achieved by precomputing embeddings every $n$ steps using the most recent network checkpoint. 
We adopt a more computationally efficient online mining strategy \cite{schroff2015facenet} 
where triplets are generated on the fly after the embeddings have been computed 
and before the evaluation of loss function and backpropagation phase.

\subsection{Evaluation methodology}

We evaluate the performance of the system by computing the following metrics:

\begin{itemize}
  \item true positive rate on pairs from the test set;
  \item top-$k$ accuracy on the test set ($k=1,5,10$).
\end{itemize}


\begin{figure}[!t]
    \centering
    \includegraphics[width=3.3in]{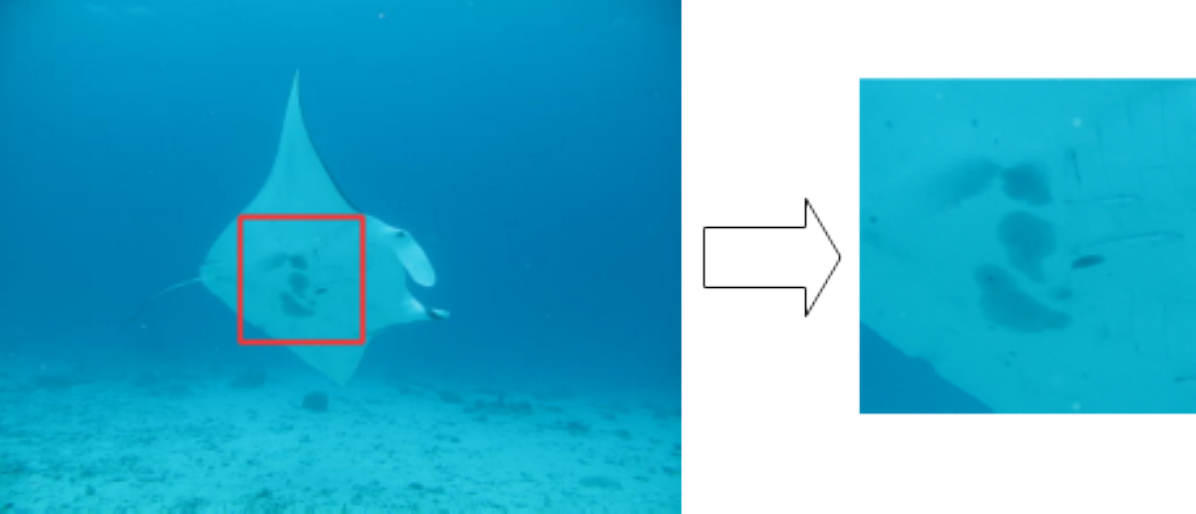}
    \caption{The user is required to draw a bounding box around the region containing the natural markings. Images are cropped to contain the pattern of interest only. This is the only input required from the user. Photo credit: Chris Garraway.}
    \label{fig:manta-cropping}
\end{figure}

\subsubsection{Validation on pairs}

The network performance is evaluated on pairs generated from the test set using a method proposed in \cite{schroff2015facenet}. The set of pairs of images from a same class is denoted as $P_{+}$ and the set of all pairs from different classes is denoted as $P_{-}$.
Let us define the set of \textit{true accepts} $\text{TA}$ for a threshold $d$ as the set of correctly classified positive pairs with a threshold $d$:

\begin{equation*}
    \text{TA}(d) \defeq \{ (i,j) \in P_{+}, \, \text{with} \, D(I_i, I_j) \leq d \}.
\end{equation*}

The set of \emph{false accepts} $\text{FA}$ is defined as the set of negative pairs that are incorrectly classified as positive with a threshold $d$:

\begin{equation*}
    \text{FA}(d) \defeq \{ (i,j) \in P_{-}, \, \text{with} \, D(I_i, I_j) \leq d \}.
\end{equation*}

We calculate the \textit{true positive rate} $ \text{TPR} $ (or \textit{recall}) and the \textit{false acceptance rate} $\text{FAR}$ for a given threshold $d$ as:

\begin{equation*}
    \text{TPR}(d) \defeq \frac{|\text{TA}(d)|}{|P_{+}|}, \quad \text{FAR}(d) \defeq \frac{|\text{FA}(d)|}{|P_{-}|}
\end{equation*}

Thanks to the relatively small size of the test datasets, all possible pairs are generated. The models are evaluated by plotting ROC curves and computing the area under the curve. 
The models are compared with respect to the true positive rate $ \text{TPR} $ at the threshold $d$ when the false 
acceptance rate $\text{FAR} = 0.01$.

\subsubsection{Accuracy evaluation on the test set} \label{sec:eval-real-life}

From a marine biologist's point of view, a reliable system should have at least 95\% top-10 accuracy. 
The accuracy of re-identification depends on the number of matching images in the database for each query image. 
We consider a realistic scenario where each query individual has two matching images in the database (our databases have at least 3 images for each individual). If there are more images per individual in the database, the task of re-identification becomes easier. 
For training, the dataset is partitioned into a training set and a test set in such a way so each individual animal appears exclusively either in the training set or in the test set.
For testing, the database is made of the training set images plus $m=2$ random images for each test individual. 
The rest of test images are used as query images.  
The accuracy is averaged over all test individuals in multiple runs by moving different images from the test set to the database. We also analyze the effect of varying the number $m$ in Section \ref{sec:experiments-number-m}. A similar evaluation procedure has been performed in \cite{review-hotspotter, manta_matcher} on different datasets.

\section{Experiments}\label{sec:experiments}

\subsection{Datasets} \label{sec:experiments-dataset}

\subsubsection{Manta ray belly patterns}
The experiments have been conducted on a dataset of manta ray images from Project Manta (a multidisciplinary research program based at the University of Queensland, Brisbane, Australia). Images have been manually checked to select the ones that show a pattern on a belly with enough clarity to be recognized by a human. See some examples in Fig.~\ref{fig:manta-dataset} (left). The dataset is challenging as it contains photos of the patterns taken at oblique angles, in a muddy water or with some small occlusions (small fish, water bubbles). 
Uninformative images 
such as the view of the back of a manta, partial views or unclear patterns have been removed from the dataset.
See examples in Fig.~\ref{fig:manta-dataset} (right). Each image has been manually annotated with a bounding box around the pattern.
Then, each image has been cropped to the area inside the bounding box (Fig.~\ref{fig:manta-cropping}). 
Manually highlighting the  belly pattern region is the only input required from the user in our application. 

The resulting dataset (see details in Table~\ref{tab:data-stats}) is partitioned into the training set (96 individuals) and the test set (24 individuals). 


\begin{figure}[!t]
    \centering
    \includegraphics[width=3.3in]{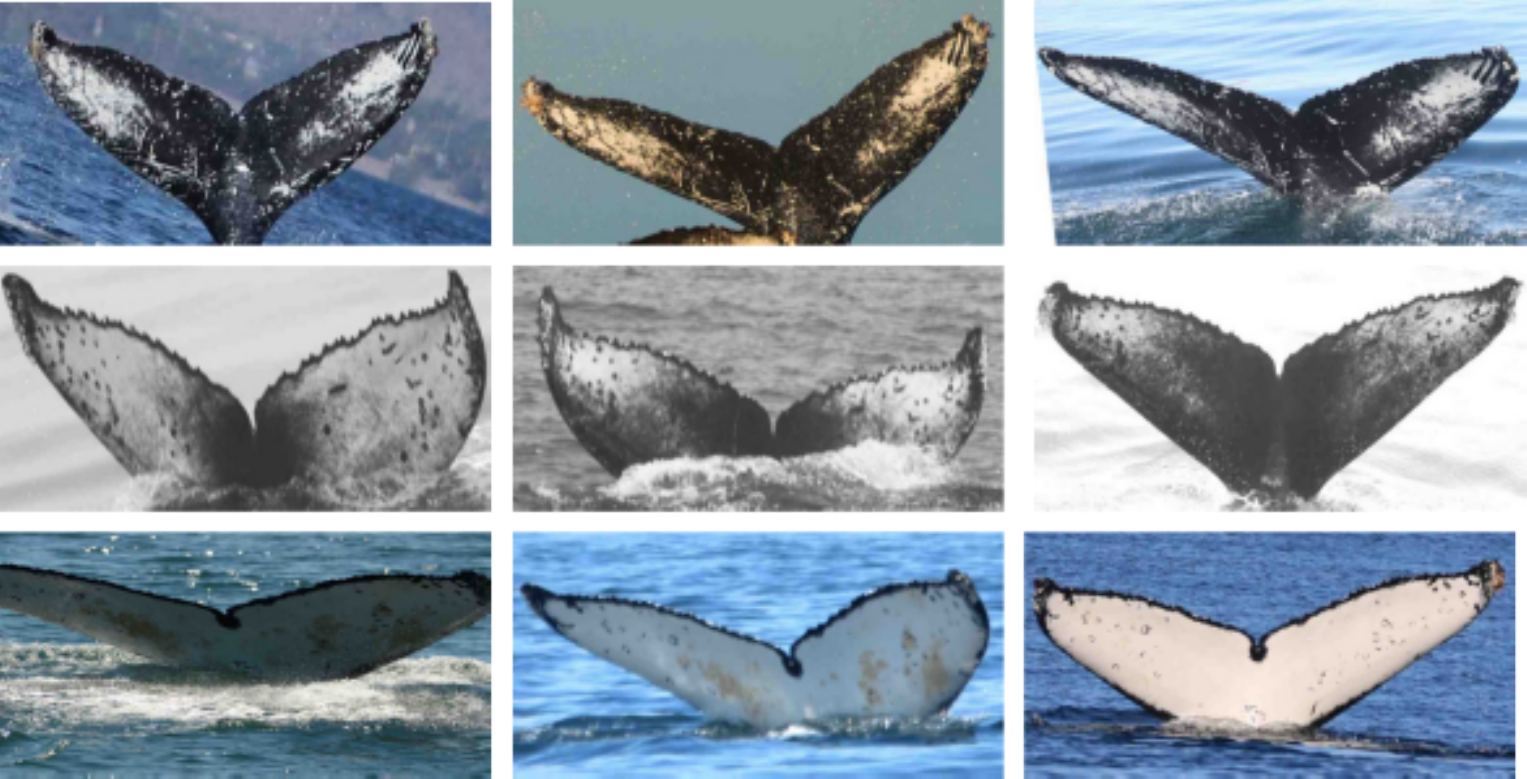}
    \caption{Images from the whale dataset. Each row shows three images of the same fluke. List of photo credits is provided in acknowledgment.}
    \label{fig:whale-dataset}
\end{figure}

\begin{table}[]
    \renewcommand{\arraystretch}{1.5}
    \caption{Statistics for the datasets}
     \label{tab:data-stats}
    \begin{tabular}{@{}lcccc@{}}
    \hline
    \multicolumn{1}{c}{} & \multicolumn{2}{c}{Manta rays} & \multicolumn{2}{c}{Whales} \\ 
    \multicolumn{1}{c}{Stats}      & Dataset       & One fold       & Dataset     & One fold     \\ \hline
    Number of images          & 1730          & $\sim$350      & 2908        & $\sim$550    \\ \hline
    Number of individuals         & 120           & 24             & 633         & 126           \\ \hline
    Average \# images per ind.  &\multicolumn{2}{c}{14}          & \multicolumn{2}{c}{5}      \\ \hline
    Min \# images per ind.      & \multicolumn{2}{c}{6}          & \multicolumn{2}{c}{3}      \\ \hline
    \end{tabular}
\end{table}

\subsubsection{Humpback whale flukes}

The dataset of humpback whales flukes comes from the Happy Whale organization (happywhale.com) \cite{happywhale2017}, \cite{happywhale2018}. 
The main challenge with this dataset is the small number of images per whale with two-thirds of the whales having one or two sightings.
For training and testing purposes we select individuals with a minimum of three images per whale resulting in a set of 2908 images for 633 unique whales, see Table~\ref{tab:data-stats}. Most of the images have already been cropped to include only the image of the fluke (see example images in Fig.~\ref{fig:whale-dataset}), however there are some noisy examples where the fluke is shown in the distance or text information 
appears at the bottom. We did not do any cropping, although this may further improve the results. 

The challenges encountered with the whale dataset are different from those of the manta ray dataset. 
Although there is no large variation in  pose or  viewpoint, 
there is a limited number of examples per individual, a combination of black-and-white and colour images, a variety of illumination conditions and some noisy images.

\subsection{Implementation details}

\subsubsection{Batch generation}
Training is performed on batches of $B \defeq P \times K$ images, where $P$ is a number of distinct classes in the batch and $K$ is a number of examples per class. During training, the whole batch is fed into the network and embeddings for the batch are computed. Embeddings are then combined into triplets based on pairwise distances according to the semi-hard triplet selection strategy discussed in Section~\ref{sec:methodology-sample-mining}. 
We use batches of 15 classes with 5 images per class for manta rays and 3 images per class for whales as this is the maximum batch size that fits into the memory of the computer utilized in these experiments.

\subsubsection{Data augmentation}
Data augmentation is used extensively during training to increase the variety in the training set. 
Transformations are applied on the fly so that at every epoch the network receives a new augmentation of the image. 
For the manta ray dataset the following geometric transformations were used: rotation up to 90 degrees, horizontal and vertical flips, 
small shifts up to 10 pixels and zooming in to 10 percent.
Most of the whale images have already a normalized view of the fluke upright. Therefore, only small rotation angles are used in data augmentation for whales.


\subsubsection{Base networks}
\label{sec:base-networks}
We compare convolutional layers of InceptionV3 \cite{inception-v2} and MobileNetV2 \cite{mobilenet-v2} as feature extractors to assess the influence of the CNN architecture on the performance of the system. One of the key differences between these two models is the number of parameters and operations. The smaller MobileNetV2 has 3.4 million parameters and 300 million multiply-adds operations \cite{inception-v2}. The bigger InceptionV3 has 23 million parameters and 5 billion multiply-adds per inference \cite{inception-v2}. 
The convolutional layers of both networks have been initialized with weights pretrained on Imagenet \cite{imagenet-ILSVRC15}.

The input size of the network depends on the  case study: the input images of whale flukes are resized to  $224 \times 448$  
because of  the shape of the region of interest; the input images for manta ray pattern 
are of shape  $300 \times 300$ for InceptionV3 and $224 \times 224$ for MobileNetV2 (pretrained weights for MobileNetV2 are available only for some input sizes). 
Images are preprocessed the same way as it was done for the model used for fine-tuning (pixel values are scaled from [0,255] to [-1,1]).

\subsubsection{Training} \label{sec:exp-train}
The Adam optimizer \cite{adam} is used for all experiments with a learning rate $10^{-5}$ and other hyperparameters with default values ($\beta_1 = 0.9$, $\beta_2 = 0.999$).
We used a learning rate $10^{-5}$ because higher values did not work well with the pretrained weights (the same has been observed in \cite{DBLP:in-defence-triplet} while training the pretrained network with the triplet loss).


In order to produce an accurate evaluation of the performance of the network, we perform k-fold cross-validation for the first experiment (Section~\ref{sec:ft-inception}).
All splits are done with respect to individuals so each individual appears only in training or test split.
Each dataset is split in five parts and five rounds of training are completed with four folds allocated for training and one fold for testing. 

All experiments have been run on a cluster with two Tesla M40 24GB GPUs and 6 CPUs.

\subsection{Performance evaluation}

\subsubsection{Fine-tuning InceptionV3 based model} \label{sec:ft-inception}


\begin{figure}[!t]
    \centering
    \includegraphics[width=3.3in]{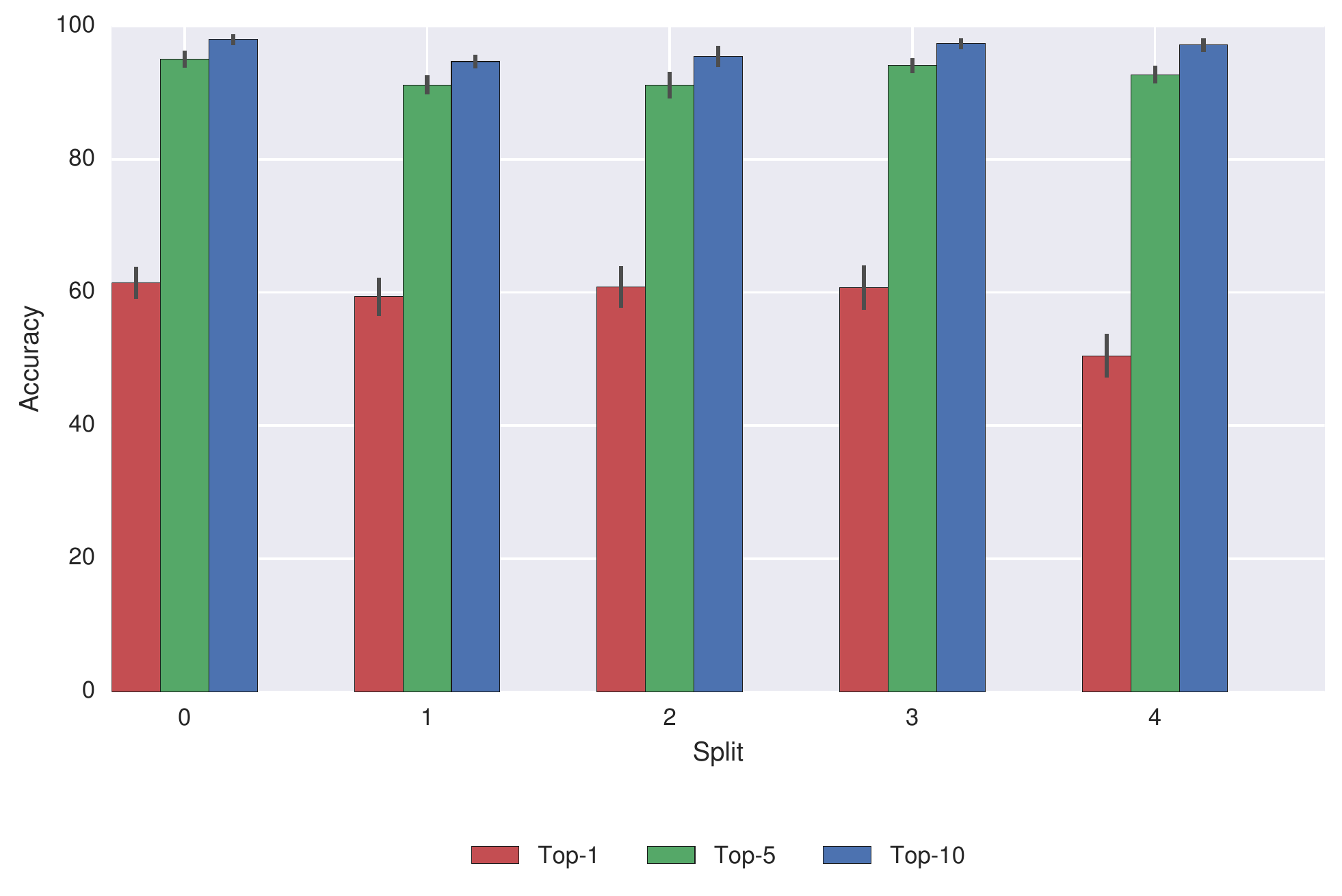}
    \caption{Top-k accuracy over 5 splits on manta ray patterns with \textit{Inception-Ft} configuration.}
    \label{fig:bar-chart-acc}
\end{figure}

\begin{table}
    \renewcommand{\arraystretch}{1.5}
    \setlength{\tabcolsep}{15pt}
    \caption{Performance of \textit{Inception-Ft} model on humpback whales and manta rays datasets separately (metrics are averaged over 5 splits)}
    \label{tab:exp-baseline}
    \centering
    \begin{tabular}{ccc}
        \hline
        \multicolumn{1}{l}{} & \multicolumn{2}{c}{Dataset} \\
        Metrics & Humpback whales & Manta rays \\
        \hline
        Top-1 & 62.78\%$\pm$1.64 & 62.05\%$\pm$3.24 \\
        \hline
        Top-5 & 88.20\%$\pm$0.67 & 93.65\%$\pm$1.83 \\
        \hline
        Top-10 & 93.46\%$\pm$0.63 &  97.03\%$\pm$1.11\\
        \hline
        TPR & 73\% & 71\% \\
        \hline
        AUC & 0.980  &  0.966 \\
        \hline
    \end{tabular}
\end{table}


\begin{figure}[!t]
    \centering
    \includegraphics[width=3.3in]{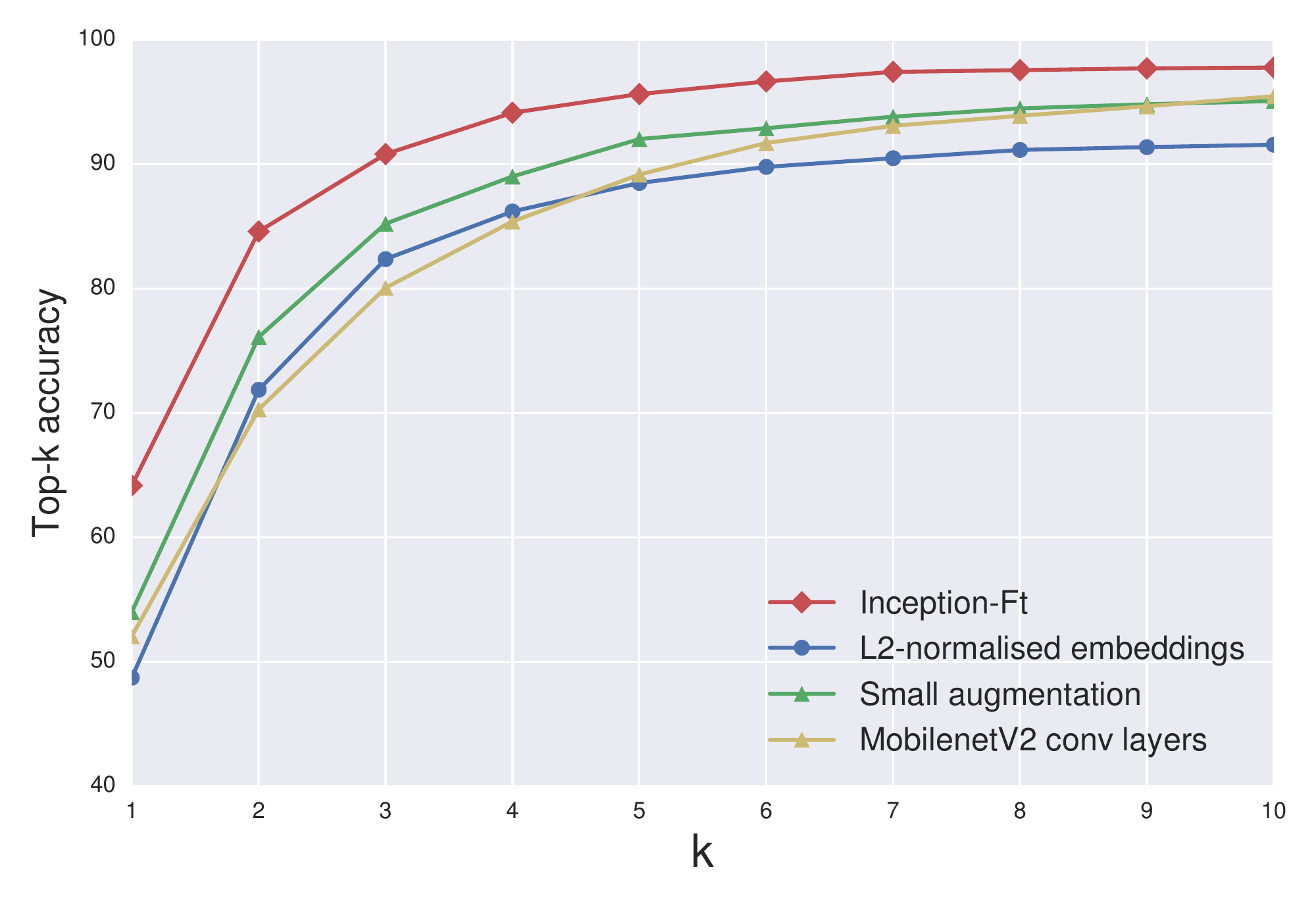}
    \caption{Top-k accuracy significantly increases at the second prediction for all configurations.}
    \label{fig:top-k-acc}
\end{figure}

We fine-tune models with InceptionV3 convolutional layers, a global pooling layer and a fully connected layer with 256 outputs on each dataset separately. We name this configuration \textit{Inception-Ft} (fine-tuned).


The metrics TPR and AUC are calculated over all possible pairs for the test fold when $\text{FAR} = 0.01$ (around 45,000 pairs for manta rays with approximately 2,000 positive pairs depending on the split; around 180,000 pairs with approximately 1,800 positive for the whales dataset). Top-k accuracy is computed for the query set where there are two matching images in the database for each query pattern. We also explore how the accuracy changes depending on the number of matches present in the database in Section~\ref{sec:experiments-number-m}.

The results of training over five splits on the manta ray dataset show that accuracy does not vary significantly over the splits, see Fig.~\ref{fig:bar-chart-acc}. 
The top-1 accuracy is $62\%$ for both datasets and the top-10 accuracy is $93\%$ for humpback whales and $97\%$ for manta rays (Table~\ref{tab:exp-baseline}).
Moreover, the graph in Fig.~\ref{fig:top-k-acc} shows that the top-k accuracy increases sharply at the second prediction ($k=2$) and top-3 accuracy is over 90\% for \textit{Inception-Ft} configuration. 

We cannot make a meaningful comparison with previous works as the results  have been reported on different datasets and the source code is not publicly available. 
Manta Matcher \cite{manta_matcher} demonstrates 50.97\% top-1 and 67.64\% top-10 accuracy on a dataset of 720 images of 265 different manta rays. 
We think that our dataset is more challenging as it contains images taken from a wider variety of angles and illumination conditions (Fig.~\ref{fig:manta-accepted}).
The best results to our knowledge for re-identification of humpback whale flukes have been reported in \cite{review-curvature}. 
The top-1 accuracy of 80\% has been achieved on a dataset of a similar size. However, the method is using integral curvature representation of the trailing edge of the flukes and is specifically designed for humpback whales. Our method is generic and not specialized for a particular species.

For the rest of the experiments we change one hyperparameter to evaluate its effect and all other parameters are kept unchanged; 
the experiments are performed on one split of manta ray dataset.

\begin{table}
    \renewcommand{\arraystretch}{1.5}
    \setlength{\tabcolsep}{15pt}
    \caption{The larger model based on InceptionV3 demonstrates better performance than the smaller model based on MobileNetV2}
    \label{tab:exp-models}
    \centering
    \begin{tabular}{ccc}
        \hline
        \multicolumn{1}{l}{} & \multicolumn{2}{c}{Base network} \\
        Metrics & MobileNetV2  & InceptionV3  \\
        \hline 
        Top-1 & 52.06\%$\pm$4.77 & 64.18\%$\pm$4.55 \\
        \hline
        Top-5 & 89.18\%$\pm$1.85 & 95.65\%$\pm$1.15 \\
        \hline
        Top-10 & 95.47\%$\pm$1.40 & 97.78\%$\pm$0.62 \\
        \hline
        TPR & 60\% & 73\% \\
        \hline
        AUC & 0.970  & 0.983  \\
        \hline
    \end{tabular}
\end{table}

\begin{table}
    \renewcommand{\arraystretch}{1.5}
    \setlength{\tabcolsep}{15pt}
    \caption{Not normalized embeddings performs better than $l_2$-normalized}
    \label{tab:exp-norm}
    \centering
    \begin{tabular}{ccc}
        \hline
        \multicolumn{1}{l}{} & \multicolumn{2}{c}{Embeddings} \\
        Metrics & $l_2$-normalized & Not normalized  \\
        \hline 
        Top-1 & 48.72\%$\pm$4.06 & 64.18\%$\pm$4.55  \\
        \hline
        Top-5 & 88.50\%$\pm$1.62 & 95.65\%$\pm$1.15  \\
        \hline
        Top-10 & 91.57\%$\pm$1.75 & 97.78\%$\pm$0.62  \\
        \hline
        TPR & 61\% & 73\%   \\
        \hline
        AUC & 0.959 & 0.983 \\
        \hline
    \end{tabular}
\end{table}

\begin{table}
    \renewcommand{\arraystretch}{1.5}
    \setlength{\tabcolsep}{6pt}
    \caption{Accuracy is not sensitive to the dimension of the embedding space}
    \label{tab:exp-embedding}
    \centering
    \begin{tabular}{cccc}
        \hline
        \multicolumn{1}{l}{} & \multicolumn{3}{c}{Embedding length} \\
        Metrics & 128 & 256 & 512 \\
        \hline 
        Top-1 & 64.46\%$\pm$3.40 & 64.18\%$\pm$4.55 & 65.75\%$\pm$4.80 \\
        \hline
        Top-5 & 95.33\%$\pm$1.08 & 95.65\%$\pm$1.15  & 94.67\%$\pm$1.61\\
        \hline
        Top-10 & 97.76\%$\pm$0.90 & 97.78\%$\pm$0.62  & 97.47\%$\pm$0.72\\
        \hline
        TPR & 72\% & 73\% & 70\%  \\
        \hline
        AUC & 0.983 & 0.983 & 0.980 \\
        \hline
    \end{tabular}
\end{table}

\begin{table}
    \renewcommand{\arraystretch}{1.5}
    \setlength{\tabcolsep}{15pt}
    \caption{Extensive augmentation (rotations up to $360^{\circ}$ and flips) of input images improve performance compared to only small amount of augmentation (rotations up to $10^{\circ}$)}
    \label{tab:exp-aug}
    \centering
    \begin{tabular}{ccc}
        \hline
        \multicolumn{1}{l}{} & \multicolumn{2}{c}{Augmentation} \\
        Metrics & Small augmentation & Extensive augmentation  \\
        \hline
        Top-1 & 54.00\%$\pm$3.32 & 64.18\%$\pm$4.55  \\
        \hline
        Top-5 & 92.03\%$\pm$1.62 & 95.65\%$\pm$1.15  \\
        \hline
        Top-10 & 95.09\%$\pm$1.28 & 97.78\%$\pm$0.62  \\
        \hline
        TPR & 58\% & 73\%   \\
        \hline
        AUC & 0.970 & 0.983 \\
        \hline
    \end{tabular}
\end{table}

\subsubsection{Influence of the base network}

We evaluate the effect of the model architecture by training two networks with convolutional layers from InceptionV3 and MobileNetV2. The larger model InceptionV3 demonstrates better performance in both validation on pairs ($\text{TPR}$ 73\% vs 60\%) and top-k accuracy (top-1 accuracy 64\% vs 52\%), see Table~\ref{tab:exp-models}. However, the difference in performance decreases for higher $k$ and top-10 accuracy is 97\% for Inception based and 95\% for MobileNet based networks.
The advantage of MobileNetV2 is a slightly faster execution but our system 
does not have to work in real-time. 
The rest of the experiments are continued with InceptionV3 convolutional layers.

\begin{figure}[!t]
    \centering
    \includegraphics[width=3.3in]{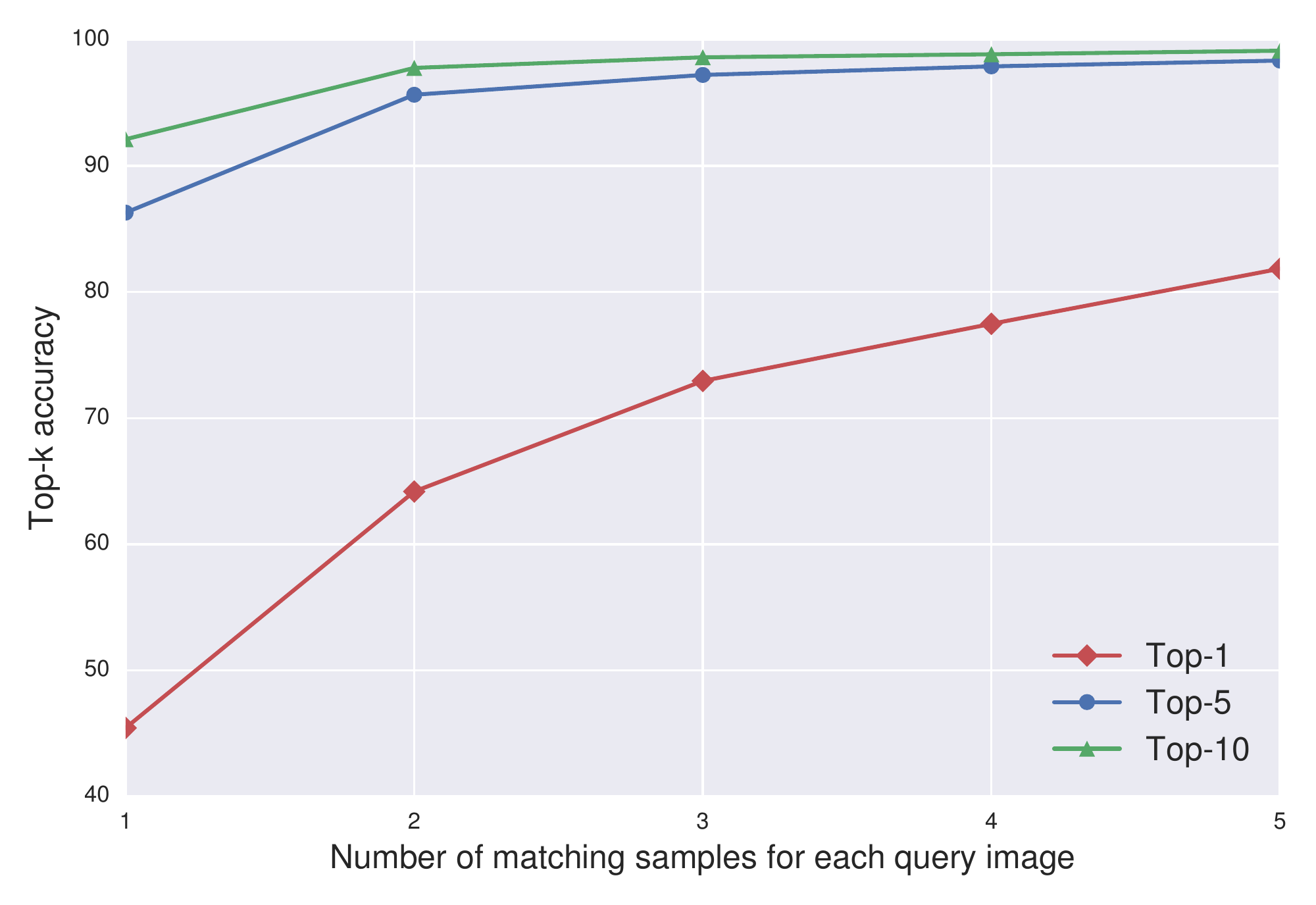}
    \caption{Accuracy of re-identification depending on the number of matching examples per query individual in the database.}
    \label{fig:top-k-m-samples}
\end{figure}


\begin{figure*}[t!]
    \centering
    \includegraphics[width=0.93\textwidth]{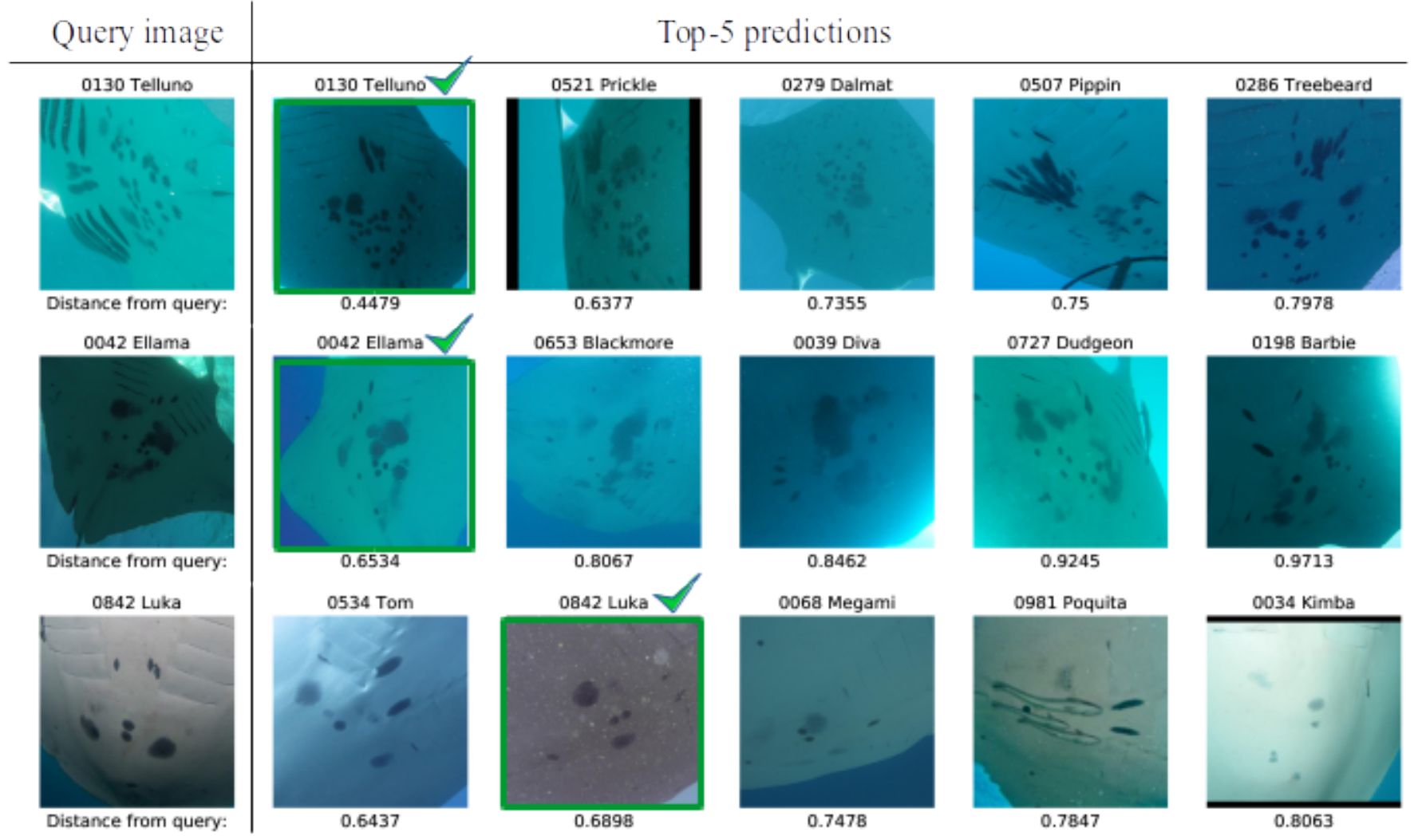}
    \caption{Three examples of correct predictions. All closest predictions share visual similarity to the query image. The pattern is correctly matched even for examples with a challenging viewpoint and illumination. Photo credits: Fabrice Jaine, Linda Earthwatch, John Lawson, Chris Garraway, Chris Garraway, Chris Kim, Maggie McNeil, Chris Garraway, Rebecca Fonskov, Kathy Townsend, Chris Dudgeon, Chris Garraway, Sarah Williamson, Ryan Jeffery, Amelia Armstrong, Ryan Jeffery, Josh Gransbury, Chris Garraway.}
    \label{fig:preds-correct}
\end{figure*}

\begin{figure*}[t!]
    \centering
    \includegraphics[width=0.93\textwidth]{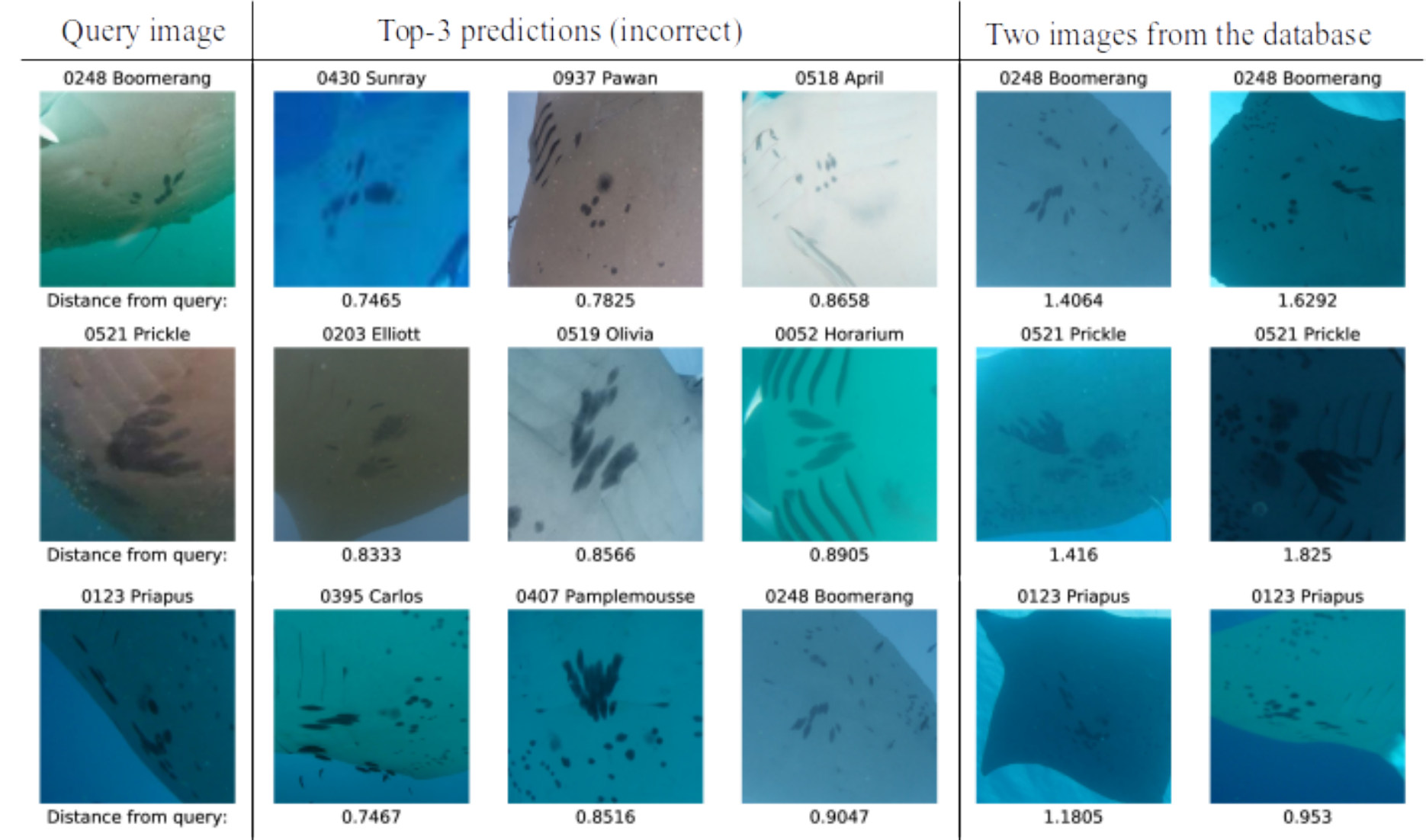}
    \caption{Three examples of incorrect matching (no match within ten closest predictions). Two images on the right are the only matching examples for each query image in the database. The query images are difficult because of the oblique angle that limits the visibility of the whole pattern. Top-3 predictions share some visual similarity to the query image. Photo credits: John Gransbury, Ian Christie, Amelia Armstrong, Mark Gray, Kathy Townsend, Kathy Townsend, Graeme Haas, Gerard Smith, Amelia Armstrong, Fabrice Jaine, Michael Rowett, Mounties Earthwatch, Mark Atkinson, Lydie Couturier, Chris Garraway, Kathy Townsend, Chris Garraway, Deg Ed.}
    \label{fig:preds-incorrect}
\end{figure*}

\subsubsection{Effect of embedding normalization} \label{sec:exp-norm}
FaceNet architecture \cite{schroff2015facenet} uses $l_2$-normalization whereas Hermans et al. \cite{DBLP:in-defence-triplet} argue that forcing the norm of the embedding to $1$  does not improve performance.
Our experiments demonstrate that restricting the embedding 
space to a hypersphere decreases the accuracy and metrics for verification on pairs. 
For example, top-1 accuracy drops from 64\% to 48\%  when we apply $l_2$-normalization, TPR decreases from 73\% to 61\% see Table~\ref{tab:exp-norm}. 
Therefore, the rest of the experiments was done without $l_2$-normalization.

\begin{figure*}[t!]
    \centering
    \includegraphics[width=0.95\textwidth]{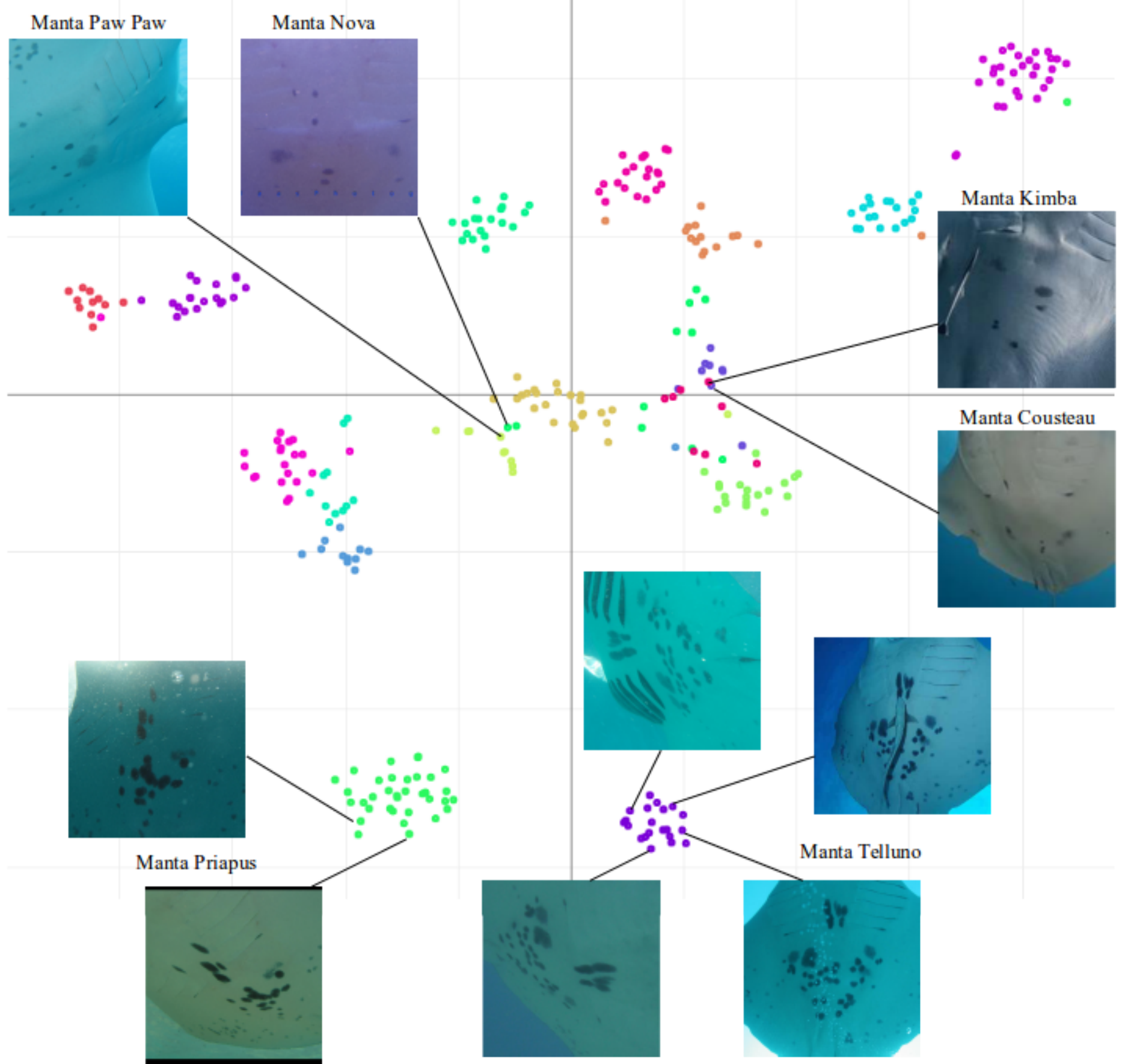}
    \caption{Visualization of embeddings computed for the manta ray test set (best viewed in colour) using t-SNE \cite{vandermaaten2008visualizing}. Embeddings for manta Telluno and manta Priapus form tight clusters and show that the learned representation is invariant to rotations, a viewing angle and small occlusions. Mixing between classes happens when the pattern has several sparse dots (manta Paw Paw and manta Nova; manta Kimba and manta Cousteau). Photo credits (in a clockwise order starting from manta Kimba: Steward Barry, Mark Gray, Fabrice Jaine, Nigel Marsh, Kathy Townsend, Lydie Couturier, Chris Garraway, Chris Garraway, Chris Garraway, Matt Prunty.}
    \label{fig:tsne-manta}
\end{figure*}


\begin{figure*}[t!]
    \centering
    \includegraphics[width=0.95\textwidth]{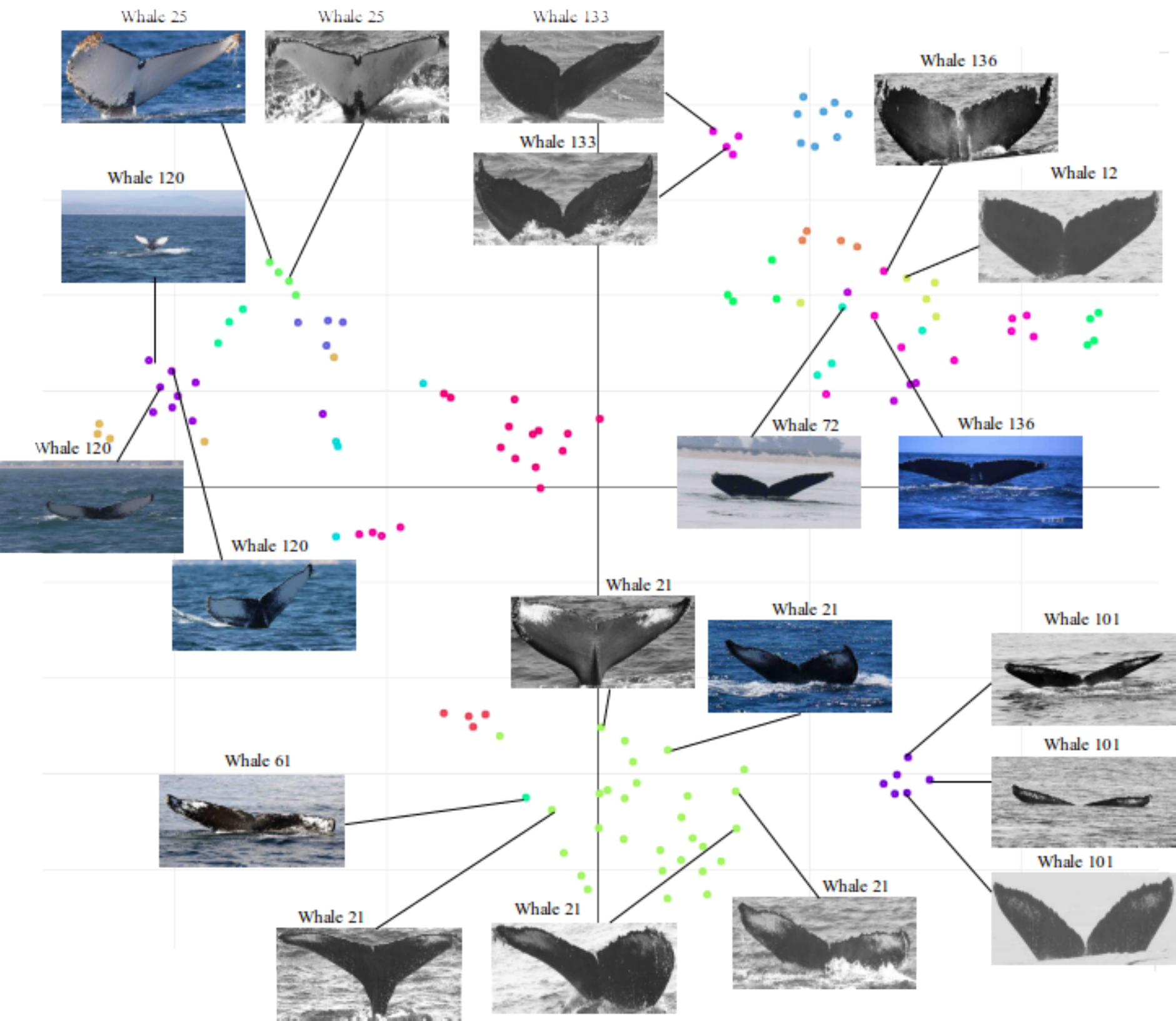}
    \caption{Visualization of embeddings computed for the test set for humpback whale flukes (best viewed in colour) using t-SNE \cite{vandermaaten2008visualizing}. Photo credit: Happywhale organization. List of photo credits is provided in acknowledgment.}
    \label{fig:tsne-whales}
\end{figure*}

\subsubsection{Influence of embedding dimension}

We tested three values for the dimension of the embedding space, 128, 256 and 512. Averaged results are reported in Table~\ref{tab:exp-embedding}. The difference between achieved accuracy is statistically insignificant and we select dimension of 256 for all other experiments.
Experiments with smaller embedding spaces (dimensions 32 and 64) showed inferior performance compared to higher dimensional spaces.

\subsubsection{Effect of data augmentation}

We have investigated the effect of data augmentation on the manta ray dataset. 
The pattern on a manta ray belly may appear at different angles so extensive data augmentation including full rotations and flips has been applied to \textit{Inception-Ft} model. 
We train the same model with rotations to only 10 degrees and no flips to estimate the influence of data augmentation on the performance. 

The experiment shows (Table~\ref{tab:exp-aug}) that the performance results of the tested model are lower when less augmentation is applied during training: top-1 accuracy drops significantly, 54\% vs 64\%, and $\text{TPR}$ has dropped to 58\% compared to 73\%. 
This demonstrates that rotations and flips of training examples facilitate learning of pattern invariance to rotations. 
However, the difference in top-5 and top-10 accuracy is less marked.

\subsubsection{Number of matching individuals in the database} \label{sec:experiments-number-m}
Previous experiments in this paper have been conducted under the condition that there are two matching images for each query individual in the database. This experiment compares accuracy for a different number of matching individuals (from one to five), see Fig.~\ref{fig:top-k-m-samples}. The fewer images in the database for each query individual, the more difficult it is for the network to find the right match. 
The number of matching examples for an individual in the database is more important for top-1  than for top-5 or top-10 accuracy. Top-1 accuracy is around 45\% for only one matching image, it increases to 64\% for two matches and reaches 81\% when there are five images in the database for each individual. Top-10 accuracy reaches 98\% with at least three images per individual in the database which is beneficial for the practical application.

\subsubsection{Visualization of predictions}
Fig.~\ref{fig:preds-correct} shows three query images and top-5 predictions of the system.
All predictions share visual similarity with a query image.
Three examples of incorrect matches alongside with top-3 predictions and two matching examples from the database are shown in Fig.~\ref{fig:preds-incorrect}. These examples are challenging as the pattern is only partly visible because of the oblique angle.

We analyze the learned representation 
with t-SNE \cite{vandermaaten2008visualizing}. 
The t-SNE algorithm  maps a high dimensional space into a two-dimensional while preserving the similarity between points. The t-SNE plot for the manta ray test set (see Fig.~\ref{fig:tsne-manta}) shows examples where embeddings for the same manta ray (manta Telluno, manta Priapus) are clustered together even when the viewpoint is different and small occlusions are present (water bubbles, small fish). Embeddings are less separated for the less distinguishable markings where a pattern consists of a small number of black marks placed sparsely (manta Paw Paw and manta Nova; manta Kimba and manta Cousteau). On the t-SNE plot for the humpback whales test set (see Fig.~\ref{fig:tsne-whales}) we observe that individuals are clustered together even when the fluke is visible from different distances (whale 120). The system is invariant to the pose of the fluke (whale 101, whale 21) and viewpoint position (whale 25). The mix between whales occurs for some totally black flukes (whales 136, 12, 72) or for the flukes with a similar colour pattern (whale 61 and whale 21).

\section{Conclusion}

We have presented a novel visual re-identification system for manta rays that is robust to viewpoint changes, variations in lighting and small occlusions. The results have been achieved by using a combination of InceptionV3 model, the semi-hard triplet mining strategy, the triplet loss function and an extensive geometric augmentation of the input images. The  practical value of the system  been demonstrated on a manta ray dataset and an humpback whale dataset. 
The system requires the user to localize the region of interest by drawing a bounding box around it. 

In the future, we plan to  further improve the system by automating the localization of the patterns of interest. 
One possible stategy is to train the network on auxiliary tasks like learning to predict the locations of specific body landmarks (tip of the wings and gills of manta rays, fluke tips and notch for whales). This would force the network to learn about the morphology of the animal. 
This ability should help induce a better representation of the spatial position of the pattern with respect to the body.


\section*{Acknowledgment}

The authors would like to thank Project Manta \\ (https://sites.google.com/site/projectmantasite/home) and Happywhale organization (happywhale.com) for the datasets of images. Photo credits for published images of humpback whale flukes: Alethea Leddy, Barry Gutradt, Casey Clark, Channel Islands NMS Naturalist Corps, Colin Garland, Dale Frink, Fernando Arcas, JB, John Calambokidis, Kate Cummings, Kate Spencer, Mark Girardeau, Richard Jackson, Ryan Lawler, Traci Phillips.

Computational resources and services used in this work were provided by the HPC and Research Support Group, Queensland University of Technology, Brisbane, Australia.

\bibliographystyle{IEEEtran}
\bibliography{biblist}

\end{document}